\documentclass[10pt,twocolumn,letterpaper]{article}

\usepackage{wacv}
\usepackage{times}
\usepackage{epsfig}
\usepackage{graphicx}
\usepackage{amsmath}
\usepackage{amssymb}

\usepackage{multirow}
\usepackage{color,xcolor,colortbl}

\newcommand{\comment}[1]{}

\renewcommand{\Re}{\mathbb{R}}

\wacvfinalcopy

\usepackage[pagebackref=true,breaklinks=true,letterpaper=true,colorlinks,bookmarks=false]{hyperref}

\ifwacvfinal\fi
\begin{document}

\title{Learning Privacy Preserving Encodings through Adversarial Training}

\author{Francesco Pittaluga\\
University of Florida\\
{\tt\small f.pittaluga@ufl.edu}
\and
Sanjeev J. Koppal\\
University of Florida\\
{\tt\small sjkoppal@ece.ufl.edu}
\and
Ayan Chakrabarti\\
Washington University in St. Louis\\
{\tt\small ayan@wustl.edu}
}

\maketitle
\ifwacvfinal\thispagestyle{empty}\fi

\begin{abstract}

  We present a framework to learn privacy-preserving encodings of images that inhibit inference of chosen private attributes, while allowing recovery of other desirable information. Rather than simply inhibiting a given fixed pre-trained estimator, our goal is that an estimator be unable to learn to accurately predict the private attributes even with knowledge of the encoding function. We use a natural adversarial optimization-based formulation for this---training the encoding function against a classifier for the private attribute, with both modeled as deep neural networks. The key contribution of our work is a stable and convergent optimization approach that is successful at learning an encoder with our desired properties---maintaining utility while  inhibiting inference of private attributes, not just within the adversarial optimization, but also by classifiers that are trained after the encoder is fixed. We adopt a rigorous experimental protocol for verification wherein classifiers are trained exhaustively till saturation on the fixed encoders.  We evaluate our approach on tasks of real-world complexity---learning high-dimensional encodings that inhibit detection of different scene categories---and find that it yields encoders that are resilient at maintaining privacy.

\end{abstract}

\begin{figure*}[!t] 
  \centering
  \includegraphics[width=\textwidth]{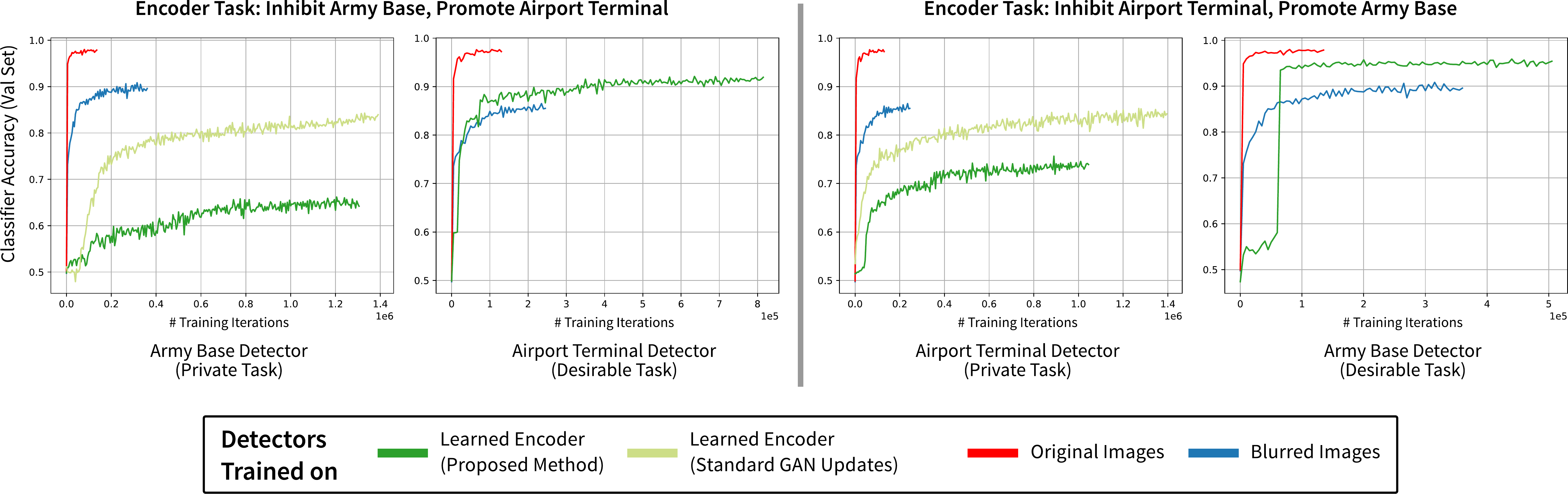}
  \caption{\textbf{Training Classifiers on Learned Privacy-preserving Encoders.} We show the evolution of validation set accuracy when learning classifiers on outputs of encoders trained to preserve privacy and maintain utility. We consider two settings: encoders trained to inhibit detection of the ``Army Base'' scene category while promoting ``Airport Terminal'', and vice-versa. For each case, we examine classifiers trained for both private and desirable tasks on encoded images. We compare to training classifiers on the original images themselves, as well as on blurred images as a naive non-task specific baseline. For private tasks, we include comparisons to encoders also trained in the same adversarial framework, but with standard GAN updates rather than our approach. Our encoders preserve information for desirable tasks, while degrading the ability to solve private tasks---both in terms of training speed and final achieved accuracy. Moreover, our approach yields encoders that are far more effective at preserving privacy than with standard GAN updates---even though the latter are able to inhibit private classifiers within adversarial training, those classifiers recover to a much greater degree once the encoders are fixed.}
  \label{fig:teaser} 
\end{figure*}

\section{Introduction}
\label{sec:intro}

Images and videos are rich in information about the environments they represent. This information can then be used to infer various environment attributes such as location, shapes and labels of objects, identities of individuals, classes of activities and actions, etc. But often, it is desirable to share data---with other individuals, un-trusted applications, over a network, etc.---without revealing values of certain attributes that a user may wish kept private. For such cases, we seek an encoding of this data that is \emph{privacy-preserving}, in that the encoded data prevents or inhibits the estimation of specific sensitive attributes, but still retains other information about the environment---information that may be useful for inference of other, desirable, attributes.

When the relationship between data and attributes can be explicitly modeled, it's possible to derive an explicit form for this encoding~\cite{dwork2008differential}. This includes the case where the goal is to encode a fixed dataset with known values of the private label (where privacy can be achieved, for example, by partitioning the dataset into subsets with different values of the private label, and explicitly transforming each set to the same value~\cite{kprivacy}). This work deals instead with the setting where the relationship between data and private attributes is not explicit, and is \emph{learned} through training an estimator.

Our goal is to find an encoding that prevents or inhibits such a trained estimator or classifier from succeeding. Note that we do not want an encoding that simply confounds a fixed classifier or estimator. Rather, we want that \emph{even after the encoding is fixed}, a classifier that has knowledge of the encoding, and which can therefore be trained on encoded training data, is unable to make accurate predictions when generalizing beyond the training set. This can be especially challenging in the vision setting when, for example, an image has potentially multiple, redundant cues towards a private environment attribute. While a specific censorship strategy could cause failures in a given estimator by interfering with the cues it depends on, given a chance to retrain, the estimator learns to use different cues still present.

To address this issue, we consider a formulation to \emph{learn} an encoding function, through adversarial training against a classifier that is simultaneously training to succeed at recovering the private attribute from encoded data. The encoder, in turn, trains to prevent this inference, while also maintaining some notion of utility---a generic objective of maintaining variance in its outputs or promoting the success of a second classifier training for a different attribute.

This is a natural formulation given the success of adversarial optimization~\cite{goodfellow2014generative}, and has in fact previously been considered in the privacy setting~\cite{edwards2015censoring, raval2017protecting,hamm2017minimax}. However, we find standard adversarial optimization to be insufficient for achieving privacy against complex inference tasks---when producing high-dimensional encodings that inhibit recovery of an image attribute whose value may be indicated by multiple redundant cues in the input, against high-capacity classifiers modeled as deep neural networks that are able to discover and exploit such cues. In these cases, we find that there is often a significant gap between the performance of the private attribute classifier within and after adversarial optimization. An encoder training simultaneously with the classifier is able to keep the latter at bay, but the classifier recovers once it is able to train against an encoder that has been fixed. We also find that training against high-capacity classifiers leads to instability in the adversarial optimization, with the encoder often converging to a trivial locally optimal solution of producing a constant output---thus achieving perfect privacy but eliminating all utility.

A key contribution of our work is thus our modified optimization approach, that implicitly regularizes the optimization by using a form of normalization in the encoder for stability, and uses modified gradient-based updates to promote learning encoding functions that \emph{permanently} limit recovery of private attributes. We also adopt a rigorous experimental protocol for evaluating privacy-preserving encoders, where classifiers are trained exhaustively till saturation on learned encoders after they have been fixed (see Fig.~\ref{fig:teaser}). We adopt this protocol to evaluate our approach on a task with real-world complexity---namely, inhibiting detection of scene categories in images from the Places-365 dataset~\cite{zhou2017places}.

\section{Related Work}
\label{sec:rw}

\noindent \textbf{Traditional Approaches to Privacy.} There exist elegant approaches to privacy that provide formal guarantees~\cite{dwork2008differential} when the relationship between data elements and sensitive attributes can be precisely characterized. This is also true for the special case when the privacy preserving task is aimed at a fixed dataset---in which case approaches such as K-anonymity~\cite{kprivacy} can be employed, since the relationship is simply the enumerated list of samples and their attribute values. Our focus in this paper, however, is on applications where this relationship is not precisely known, and data elements to be censored are high-dimensional and contain multiple, redundant cues towards the private label that a learned estimator could be trained to exploit. 

Much prior work on achieving privacy with such data, especially with images and videos, has relied on domain knowledge and hand-crafted approaches---such as pixelation, blurring, face/object replacement, etc.---to degrade sensitive information~\cite{fd5,Boyle01,fd8,fd4,fd2,fd6}. These methods can be effective in many practical settings when it is clear what to censor, and some variants are even able to make the resulting image look natural and possess chosen attributes---e.g., replacing faces with generated ones~\cite{brkic2017know,di2017gp,meden2018k} of different individuals with the same expression, pose, etc. However, we consider the general case when all cues in an image towards the private attribute can not be enumerated, and that an adversary seeking to recover that attribute will learn an estimator specifically for our encoding. This makes the goal of learning an encoding significantly more challenging, since modern classifiers, such as those based on deep neural networks, are able to learn to make accurate predictions even from severely degraded data (e.g., \cite{igor}).

\noindent \textbf{Adversarial Training.} This naturally motivates an adversarial framework for \emph{training} an encoding function, against a classifier being trained simultaneously trained to predict the private attribute. Our approach builds on the recent success of adversarial training for learning generative adversarial networks (GANs)~\cite{goodfellow2014generative}, which demonstrated the feasibility of using stochastic gradient descent (SGD) to optimize a min-max objective involving deep neural networks with competing goals. While the theoretical stationary point of such optimization is where either network can not improve when the other is fixed, such a point is rarely achieved (or even sought) in practice when training GANs. In stark contrast, it is critical in our setting for the encoder to reach a point where it maintains its success even after it is fixed, while its adversary continues to train. Also worth noting are recent works on adversarial examples~\cite{fool0,fool,fool2,fool3} that learn perturbations to cause incorrect predictions. But, these are trained as attacks against fixed classifiers that expect natural images, and such classifiers recover when re-trained on examples with the perturbations.

\noindent \textbf{Domain Confusion.} There are interesting similarities between our formulation and those of various domain adaptation / confusion methods \cite{coral, tzeng2015simultaneous, ganin2016domain, ghifary2016deep, tzeng2017adversarial}. Some of these also set up an optimization problem to derive a feature representation that is less indicative of a specific label (a private label for us, domain identity for them). However, while domain confusion approaches can be thought of as optimizing a similar objective function, they have a fundamentally different goal: generalization across domains, rather than preventing information leakage to a determined adversary. Domain confusion methods seek to ensure that classifiers trained on their learned features transfer across domains. To achieve this, it suffices to train against relatively simple domain classifiers, whose actual accuracy need only be inhibited during adversarial training. In contrast, we evaluate our encoder against much deeper classifiers as the adversary, and measure success by allowing this classifier to train \emph{after} the encoder has been fixed. Ours is thus a substantially different setting, which requires innovations in how the optimization is carried out. 

\noindent \textbf{Adversarial Privacy.} The closest formulations to ours are those of \cite{edwards2015censoring,raval2017protecting,hamm2017minimax}. These techniques also employ different forms of adversarial optimization to learn image transformations that will prevent a classifier (trained on transformed images) from solving some sensitive task. While these methods provide an interesting proof of concept, they target relatively simple private tasks---namely preventing the detection of synthetically superimposed text~\cite{edwards2015censoring} or QR codes~\cite{raval2017protecting}, and show that adversarial training learns to detect and blur the relevant regions---or attempt to censor low-dimensional feature vectors~\cite{hamm2017minimax}.

As our experiments show, directly applying traditional adversarial optimization---while successful in domain adaptation~\cite{tzeng2015simultaneous, ganin2016domain, ghifary2016deep, tzeng2017adversarial} and the more limited privacy tasks~\cite{edwards2015censoring,raval2017protecting,hamm2017minimax}---fails when trying to persistently inhibit  powerful classifiers trained for complex real-world tasks on high-dimensional encodings of natural images (see Fig.~\ref{fig:teaser}). This work proposes an approach that is able to solve the underlying practical challenges, and by doing so, opens up the possibility of using adversarial optimization  practically and effectively for an important new application: privacy.

\section{Learning Private Encoding Functions}
\label{sec:contrib}

In this section, we begin by describing the formulation for an adversarial framework to train an encoder to inhibit classifiers for chosen sensitive attributes. This takes the form of optimizing a min-max objective---similar to those used in traditional GANs~\cite{goodfellow2014generative},  adversarial domain adaptation~\cite{tzeng2015simultaneous, ganin2016domain, ghifary2016deep, tzeng2017adversarial} and the recent works on adversarial privacy~\cite{edwards2015censoring,raval2017protecting,hamm2017minimax}). We analyze this formulation, and discuss ways to incorporate different types of constraints to maintain utility. We then describe our optimization approach, that promotes stability during training and strengthens the learned encoder's ability to maintain privacy.

\subsection{Privacy as Adversarial Objective}

We consider the following setting: when training the encoder, we have a training set labeled with values of the private attribute. Once the encoder has been trained, we seek to limit the ability of an adversary, with knowledge of this encoding function, to train an estimator for the private attribute. This means that after the encoder is fixed, we assume the adversary is able to train an estimator on an \emph{encoded} labeled training set (by applying the encoding function on a regular training set), and we seek to restrict the performance of this estimator on encoded validation and test sets. Note that we do not seek to prevent the estimator from performing well on the training set itself, e.g., through memorization. Our goal is to limit generalization accuracy.

We let $E: \Re^N \rightarrow \Re^{N'}$ denote our encoding function that maps an image $x \in \Re^N$ to an encoded counterpart $x' = E(x) \in \Re^{N'}$,  with the goal of preventing the estimation of a private attribute $u(x) \in \mathbb{U}$ from the encoded image $x'$. Consider a parameterized estimator $\Phi(x';\theta_u)$ with learnable parameters $\theta_u$ that produces an estimate $\hat{u}$ of $u(x)$ from the encoded image $x'$. Then, given a loss $L(\hat{u},u):\mathbb{U}\times\mathbb{U}\rightarrow R$, our desired encoding function is $E=\arg \min_E~~I(E;u)$ where
\begin{equation}
  \label{eq:loss}
  I(E;u) = -\min_{\theta_U}~~\underset{p(x)}{\mathbb{E}}~~L\left(\Phi\left(E(x);\theta_U\right),u(x)\right).
\end{equation}

\noindent \textbf{Theoretical Analysis.} Note that this is a min-max optimization between the parameters of the encoder $E$ and estimator $\Phi$. Consider the case when $\mathbb{U}$ is a discrete label set, $\Phi$ is a classifier that produces a probability distribution over these labels, and $L$ is the cross-entropy loss. Given an encoder $E$, let $p_E(x')$ denote the distribution of encoder outputs, and $p_{E}(x'; u)$ the distribution of encoder outputs $x'$ conditioned on the label being $u$. Further, let $\pi_u$ be probability of label $u$ (i.e., $\int_{u(x)=u} p(x) dx$), then $p_E(x') = \sum_u \pi_u p_{E}(x'; u)$. Following the derivations in \cite{goodfellow2014generative}, since the optimal output of a classifier for label $u$ is: $\Phi(x')_u = \pi_u p_{E}(x'; u) / p_{E}(x')$, it follows that:
\begin{align}
  \label{eq:mutual}
  &I(E;u)=\int p(x)\log[\pi_{u(x)}p_{E}(E(x);u(x))/p_{E}(E(x))]dx \notag\\
  &=  \sum_u \pi_u\log \pi_u - \int_{x'}p_E(x')\log p_E(x')~dx'\notag\\&~~~~ + \sum_u\pi_u \left(\int_{x'}p_E(x'; u)\log p_E(x';u)~dx'\right)\notag\\&= -H(U) + h(X') - h(X'|U),
\end{align}
where $h(X')$ is the differential entropy of encoder outputs $x'$, $h(X'|U)$ is their conditional entropy given label $u'$, and $H(U)$ is entropy of the label distribution. Therefore, $I(E;u)$ is the mutual information between encoded outputs and the private label $u$ up to a constant ($-H(U)$). Note when $u$ is a binary label and both classes are equally balanced ($\pi_0=\pi_1$),  $I$ is also the Jensen Shannon divergence between the two class distributions of encoder outputs.

\vspace{1em}\noindent \textbf{Maintaining Utility.} Absent any other constraints, $I(E;u)$ is trivially minimized by an encoder $E(x) = C$ that outputs a constant independent of the input. While such an encoding would indeed achieve absolute privacy, it would be useless for all other tasks as well. We discuss two constraints to maintain utility: a generic one in terms of variance, and one with an objective of promoting specific desirable tasks.

For the generic constraint, we require that, on average (over samples of data $x$), each element of the encoded output have zero mean and unit variance, i.e., $\mathbb{E} \  E(x)_i = 0$ and $\mathbb{E} \  E(x)_i^2 = 1, \forall i \in \{1\ldots N'\}$, where $E(x)_i$ denotes the $i^{th}$ element of $x'=E(x)$. Therefore, the encoder is constrained to produce outputs with reasonable diversity, even as it tries to remove information regarding the label $u$. This constraint is aimed at maintaining information content in the encoded outputs, so that these outputs may be informative for estimating attributes other than $u$.

Our second formulation for maintaining utility is defined in terms of allowing the recovery of one or more specific \emph{desirable} attributes. Specifically, for such an attribute $v(x)$, we can define a corresponding $I(E;v)$ similar to \eqref{eq:loss}:
\begin{equation}
  \label{eq:loss2}
  I(E;v) = -\min_{\theta_V}~~\underset{p(x)}{\mathbb{E}}~~L\left(\Phi\left(E(x);\theta_V\right),v(x)\right).
\end{equation}
An encoder to preserve $v$ while inhibiting $u$ is given by:
\begin{equation}
  \label{eq:encdef2}
  E = \arg \min  I_u(E) - \alpha I_v(E),
\end{equation}
where $\alpha>0$ is a scalar weight. Note that we enforce the zero mean and unit variance constraints on the encoder outputs for this case as well. The objective above involves an adversarial optimization with a \emph{collaboration} between the encoder $E$ and desirable attribute classifier parameters $\theta_v$, against the classifier parameters $\theta_u$ for the private attribute. 

\vspace{1em}\noindent \textbf{Stability.} In contrast to the standard GAN setting, where the generator seeks to produce outputs to match a fixed data distribution, the encoder in our setting has control over all the conditional distributions that it is trying to bring close. This has consequences on the \emph{stability} of the optimization process. While GANs are affected by degeneracy caused by the discriminator reaching perfect accuracy, \eqref{eq:loss} is plagued by a different form of instability. In our setting, optimization is prone to collapse to the trivial solution of $E(x)=C$, despite this being a violation of the variance constraint. This occurs when gradient-based updates lead to internal layers of the encoder being stuck at saturation, at which point the encoder stops updating. As described in the next section, our approach to optimization includes a form of normalization on the encoder to address this instability.

\vspace{1em}\noindent \textbf{Encoder and Classifier Architecture.} We use deep-neural networks to model the estimators $\Phi(\cdot;\theta_u)$ and $\Phi(\cdot;\theta_v)$. In particular, we consider binary attributes and $\Phi$ corresponds to a classifier trained with a cross-entropy loss $L$. We also use a deep neural network to model the encoder $E$, and the minimization in \eqref{eq:loss} and \eqref{eq:loss2} are over network weights given a chosen architecture. In this work, we focus on the case when the inputs $x$ are images, and the encoder also produces image-shaped outputs ($224\times 224$ RGB images as input, and three channel $28\times 28$ encoded outputs). Here, we use convolutional layers and spatial pooling layers in both the classifier and encoder architectures. The encoder's final layer is followed by a \emph{tanh} non-linearity to produce outputs that saturate between $[-1.0,1.0]$. We enforce the zero mean and unit variance constraints on the encoder outputs by simply placing a batch-normalization layer~\cite{bnorm} after the output of the final layer (in practice, we put this layer prior to the pre-\emph{tanh} activation), without any learnable scaling or bias.

\begin{figure*}[!t]
\begin{center}
  \includegraphics[width=0.85\textwidth]{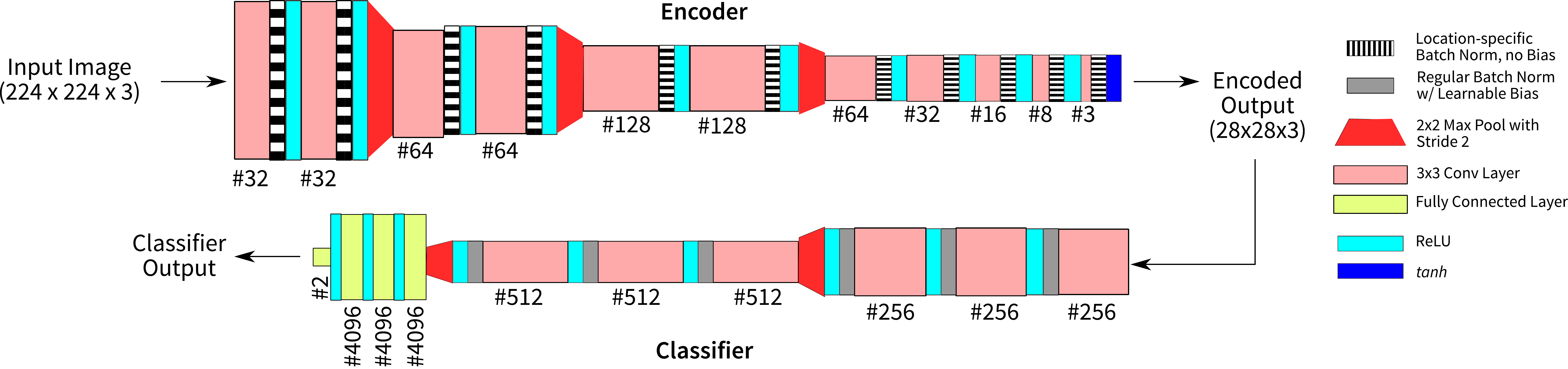}\vspace{-1em}
\end{center}
  \caption{\textbf{Encoder and Classifier Architectures}. We use convolutional architectures for both the encoder and classifier networks. To ensure stable optimization and prevent the encoder from collapsing to a trivial solution, we use per-location normalization (depicted above in black-and-white bars) without  biases after every convolution layer in the encoder.}
\label{fig:arch}
\end{figure*}

\subsection{Optimization Method}
\label{sec:opt}

We train the encoder to minimize $I(E;u)$ by applying alternating gradient updates to the encoder and classifier(s). These gradients are computed on mini-batches of training data for the classification tasks, with  different batches used to update the encoder and classifiers. The classifiers take gradient steps to minimize their losses with respect to the true labels of the encoded data. When optimizing with the desirable attribute classifier, the encoder's update is also based on minimizing that classifier's loss.

\vspace{1em}\noindent \textbf{Updates with Label Flip Loss.} When computing gradients for the encoder with respect to the private attribute classifier, we find the negative cross-entropy as indicated in \eqref{eq:loss}, as well as the log-loss with respect to the incorrect label typically used in GAN training~\cite{goodfellow2014generative} to be insufficient for our setting. Note that both these losses encourage the encoder to drive the classifier to make mis-classifications with high confidence---based on the current state of the classifier. However, the classifier can easily recover from such a state after the encoder is fixed, by simply reversing its outputs (true for false, and vice-versa)---because if the classifier has high-confidence but incorrect predictions, this still implies a separation in the per-class distributions of encoder outputs.

We propose a modified loss for encoder updates that provides a more direct path to minimizing the mutual information towards the private label: we compute gradients with respect to the cross-entropy loss treating the \emph{opposite of whichever label the classifier currently predicts} as the true label. Thus while the classifier itself trains to increase its accuracy, the encoder seeks to reduce the classifier's confidence in its predictions, \emph{be they correct or not}. We find that this approach typically causes the encoder to have a more permanent effect on private classification ability that persists even after the encoder has been fixed, as demonstrated by our experiments in the next section.

\vspace{1em}\noindent \textbf{Stability with Normalization.} As discussed previously, a significant source of instability in our setting is the encoder collapsing to the  trivial solution, where it produces a constant output independent of the input. As we will illustrate with experiments, without any further constraints, training frequently collapses to this degenerate solution despite the normalization constraint at the output enforced by a batch-normalization layer. This is caused by collapse in the intermediate layers, that are driven to producing constant outputs--either by the kernel weights going to 0, or biases to large negative values that saturate the ReLUs---and once this happens, the gradients to the encoder vanish and training is unable to move away from this solution.

To address this, we include normalization at the output of every layer in the encoder network to make the activations have zero mean and unit variance. Specifically, we add a normalization layer  after every convolution and completely remove all learnable biases (ensuring that half of all ReLUs are activated). However, we find using standard batch-normalization, which normalizes activations both across the batch and all spatial locations,  to be insufficient. This is because even with such normalization, the encoder has the ability to produce a constant output---which has different values at different spatial locations to satisfy the variance constraint, but has the same value for a given location for all inputs. (Although the encoder is convolutional, it is able to achieve this by detecting padded values at the border).

Thus, even when using standard batch-normalization after every layer and removing all learnable biases, encoder training remains unstable and often collapses to the trivial solution. Therefore, we use a ``per-location'' normalization layer that separately normalizes the activation at each spatial location, with statistics of each location computed by averaging over a batch (this is equivalent to treating the output of convolutional layers as a single large vector). This layer thus forces the encoder to produce different outputs for different images, and we use this per-location normalization at all layers, including the output. As our experiments show, this strategy reliably prevents collapse to the trivial solution, and leads to stable training in every experiment across a wide range of tasks and settings.

\section{Experimental Results}
\label{sec:exp}

\subsection{Preliminaries}

\noindent \textbf{Tasks and Dataset.} We evaluate our framework on its ability to inhibit identification of specific scene categories on images from the Places-365 dataset~\cite{zhou2017places}. Identification is framed as binary classification: whether an image belongs to a specific category or not. We train different encoders to inhibit detection of a specific category. We then fix each encoder, and evaluate privacy as the ability to train a classifier for that category, and utility as the ability to train classifiers for \emph{other} categories. For each identification task, we train and evaluate classifiers on a balanced dataset where half the images belong to that category---therefore, the ``prior'' for each task is chance. These sets are constructed from two groups of ten categories each, from Places-365\footnote{\textbf{Group 1}: {arch, army base, airport terminal, airfield, alley, arena hockey, amusement park, apartment bldg., aquarium, arena rodeo}. \textbf{Group 2:} {amphitheater, auto showroom, airplane cabin, arch.\ excavation, art studio, artists loft, assembly line, athletic field, atrium, auto factory}.}. The negative examples for each identification task are uniformly sampled from the other nine categories in the same group. We construct non-overlapping training, validation, and testing sets from the official Places-365 training set. 

Inputs to our encoder are RGB images of a fixed size $224\times 224$. These were constructed from the Places-365 images---with random scaling and crops for data augmentation during training, and a fixed scale and center-crop for evaluation. The encoder produces $28\times 28$ three-channel images as output, and these are provided as input to the classification networks. The architectures of both the encoder and classification networks (we used the same architecture for all tasks) are shown in Fig.~\ref{fig:arch}.

\vspace{1em}\noindent \textbf{Training Details.} We train various encoders to inhibit different identification tasks as described in Sec.~\ref{sec:contrib}, some with generic variance constraints and others with the objective of promoting specific desirable tasks as in \eqref{eq:encdef2}(with $\alpha=2^{-4}$). We use the Adam optimizer~\cite{adam}, and due to our dependence on per-location normalization, train with a large batch size of 128 images. We begin by training the classifier for 5k iterations as ``warm up'' against a randomly initialized encoder, and then proceed with alternating updates to the encoder and classifiers. The learning rate for the classifier is kept fixed at $10^{-4}$. For the encoder, we begin with a rate of $10^{-4}$, but then drop it by $(0.1)^{1/4}$ every 200k iterations. Empirically, dropping the learning rate has a significant effect on subsequent classification performance after the encoder is fixed, since the encoder now trains to inhibit a classifier that is able to adapt at increasingly faster relative rates. We train the encoders for a total of 860k iterations.

\definecolor{gry1}{gray}{0.9}
\definecolor{gry2}{gray}{1.0}
\definecolor{gry3}{gray}{0.98}
\definecolor{nbak}{rgb}{1.0,0.9,0.9}
\definecolor{pbak}{rgb}{0.9,1.0,0.9}
\newcommand{\rodd}{\rowcolor{gry2}}
\newcommand{\reven}{\rowcolor{gry1}}
\newcommand{\rdd}{\rowcolor{gry3}}

\vspace{1em}\noindent \textbf{Verification Protocol.} After training, we fix the encoders and measure success at achieving privacy based on limiting the ability of a classifier to learn to solve the negative task, and utility based on solving non-private tasks. To evaluate this, we train classifiers from scratch, for each task and each encoder, using encoded images for training. We train all classifiers also using Adam, with the initial learning rate set to $10^{-5}$ (we empirically found higher values to be unstable) in all cases except for those trained on the original images---where we were able to use a higher initial learning rate of $10^{-4}$. We keep training all classifiers at this learning rate till the validation accuracy saturates, and follow this one learning rate drop by a factor of 0.1 and continue training till the accuracy saturates again. Note that for private tasks against encoders learned using the proposed framework, this often requires training classifiers for \emph{orders of magnitude more iterations} than on original images.

\subsection{Comparison to Traditional Optimization}

\begin{table}[!t]
  \renewcommand{\arraystretch}{1.3}
  \begin{center}
  \begin{tabular}{|c||c|c|}
    \hline
    Task & -Army Base & -Airport T.\\
         & (+Airport T.) & (+Army Base)\\\hline\hline
    \multicolumn{3}{|l|}{\bf \scshape Within Adversarial Optimization: Val Set}\\\hline
    \reven {Standard GAN} & \bf 51.1\% & \bf 50.1\%\\\hline
    \rdd {Label Flip} & 52.1\% & 53.6\%\\\hline\hline
    \multicolumn{3}{|l|}{\bf \scshape Verification w/ Fixed Encoder: Val Set}\\\hline
    \reven {Standard GAN} & 83.9\% & 84.4\%\\\hline
    \rdd {Label Flip} & \bf 64.2\% & \bf 73.9\%\\\hline\hline
    \multicolumn{3}{|l|}{\bf \scshape Verification w/ Fixed Encoder: Test Set}\\\hline
    \reven {Standard GAN} & 84.3\% & 84.5\%\\\hline
    \rdd {Label Flip} & \bf 65.3\% & \bf 73.5\%\\\hline
  \end{tabular}
  \end{center}
  \caption{Private Attribute Classifier performance on encoders trained with flipped-label vs standard GAN updates: (a) within adversarial optimization; and (b) on verification with fixed encoders.}
  \label{tab:ablat1}
\end{table}

\newcommand{\cpos}{\cellcolor{pbak}}
\newcommand{\cneg}{\cellcolor{nbak}}

\begin{table*}
\begin{center}
\begin{tabular}{ |l|cccccccc| } 
\cline{2-9}
\multicolumn{1}{l|}{} & \multicolumn{8}{|c|}{ \textbf{\scshape Binary Scene Classification Performance (Chance = 0.5)}} \\
\hline
\multirow{2}{*}{Encoder} & 
{\small Army} & \small Airport & \multirow{2}{*}{\small Airfield} & \multirow{2}{*}{\small Arch} & \multirow{2}{*}{\small Alley} & \small Hockey & \small $\dagger$Amphi-  & \small$\dagger$Auto \\ 
&Base& \small Terminal &&&& \small Arena & \small theater & \small Showroom \\  
\hline
\rodd \small Identity   &\cpos .983&\cpos .967&\cpos .982&\cpos .942&\cpos .969&\cpos .995&\cpos .951&\cpos .958 \\ 
\reven \small Naive Blur &\cneg .914&\cneg .869&\cneg .951&\cneg .782&\cneg .881&\cneg .966&\cneg .843&\cneg .898 \\ 
\hline
\multicolumn{9}{l}{\vspace{-.5em}}\\
\hline
\reven \small -Army Base        &\cneg.754&.761&.896&.678&.781&.919&.803&.827 \\
\rodd \small -Airport T. &.796&\cneg .750&.891&.694&.832&.915&.815&.851 \\ 
\reven \small -Airfield         &.639&.667&\cneg .811&.613&.712&.824&.701&.565 \\ 
\rodd \small -Arch             &.796&.806&.905&\cneg.701&.848&.922&.826&.868 \\ 
\hline
\multicolumn{9}{l}{\vspace{-.5em}}\\
\hline     
\rodd   \parbox[c]{5em}{\vspace{.075cm}\small +Army Base\\-Airport T.\vspace{.075cm}} 
                        &\cpos.956 ~&\cneg .736&.873 &.621&.774&.877&.789&.620  \\ 
\reven \parbox[c]{5em}{\vspace{.075cm}\small -Army Base\\+Airport T.\vspace{.075cm}} 
                         &\cneg.653 ~&\cpos .916&.836&.612&.751&.821&.785&.721  \\ 
 \rodd \parbox[c]{5em}{\vspace{.075cm}\small -Army Base\\+Airport~T.,\\~~~~~~~~~~Alley\vspace{.075cm}} 
                         &\cneg.717&\cpos.930&.897&.723&\cpos.912&.921&.813&.826  \\           
 \reven \parbox[c]{6em}{\vspace{.075cm}\small -Army Base\\+Airport~T.,\\~~~~~~~~~~Airfield, Alley\vspace{.075cm}} 
                         &\cneg.807&\cpos.939&\cpos.967&.741&\cpos.890&.918&.862&.890  \\                         
\rodd  \parbox[c]{5em}{\vspace{.075cm}\small -Army Base\\+Arch\vspace{.075cm}} 
                         &\cneg.801~&.802&.932&\cpos.867&.840&.938&.870&.855  \\                            
\hline
\end{tabular}
\end{center}\vspace{-1em}
~{\scriptsize \hspace{51em} $\dagger$ Categories from Group 2.}\vspace{1em}
\caption{Scene Category Detection Performance with Different Encoders. We consider a variety of encoders (one per row) using our approach to inhibit different tasks, with both generic output variance constraints (rows 3 to 6), as well as to promote specific desirable tasks (rows 7 to 12). For comparison, we also consider the original images themselves (row 1), as well as images degraded by blur as a simple baseline (row 2). For each encoder, we train classifiers on encoded images to solve different scene detection tasks---including private tasks that the encoders were trained to inhibit (red), desirable tasks they were trained to promote (green), and tasks that the encoder was trained neither to inhibit nor promote (rest)---and report test set classification accuracy. Note that every classifier (one for each cell in the table) is trained exhaustively till saturation against a fixed encoder. For private task classifiers training on encodings produced by our learned encoders, this takes between \textbf{1-4.5M} iterations, while classifiers training on the original images train much faster (all crossing 90\% validation accuracy within \textbf{20k} iterations.).}\label{tab:results}
\end{table*}

\noindent \textbf{Effect of  Label Flip-based Updates.} We first illustrate the importance of computing gradient updates for the encoder with respect to flipped labels as described in Sec.~\ref{sec:opt}, instead of standard updates---used in GAN training, domain adaptation, etc.---that minimize loss with respect to the incorrect label, to inhibit the private attribute classifier. In Table~\ref{tab:ablat1}, we report results for two cases of training encoders with respect to the objective in \eqref{eq:encdef2} to inhibit detection of one scene category while promoting detection of another---alternating between ``Army Base'' and ``Airport Terminal'' as the private and useful tasks. We report accuracy of classifiers for the private attributes at the end of adversarial encoder training. Then, we measure the accuracy of classifiers trained exhaustively under our verification protocol after fixing the encoders, and report these on encoders trained both with standard GAN updates and our flipped label loss.

We see that within adversarial training, the private classifiers are nearly perfectly inhibited and have accuracy close to chance---indeed, the standard GAN updates even lead to slightly better inhibition. But crucially, once the classifiers are allowed to train while the encoder remains fixed, they recover to a significantly greater degree against encoders trained with the GAN updates (this is also visualized in Fig.~\ref{fig:teaser}) than against those trained using our approach. This highlights the importance of the flipped label loss in learning a resilient encoder that has a permanently inhibiting effect on private classifier performance.

\vspace{1em}\noindent \textbf{Stability with Per-location Normalization.} Next, we discuss the need to stabilize training by using per-location normalization without biases after every intermediate layer in the encoder---note that all encoders in Table.~\ref{tab:ablat1} were trained with this normalization. As described in Sec.~\ref{sec:contrib}, the encoder in our setting has access to all inputs to the classifier, and thus collapse to a degenerate solution with constant outputs is a concern. To demonstrate the importance of normalization, we train encoders for inhibiting ``Army Base'' (this time with a generic variance constraint) with the following modified settings: no batch normalization in intermediate layers, regular batch-normalization (with spatial and batch averaging) without biases, per-location normalization but with biases, and per-location normalization without biases (the proposed setting). In the first case, the encoder collapses to a state where the output variance constraint is violated for all outputs. In the next two cases, we see a partial collapse in the solution. With regular batch normalization without biases, 75\% of the final outputs have zero variance. And while per-location batch normalization with biases is able to prevent any of the output variances from going to zero, 90\% of them have variance less than 0.5. In contrast, our approach of per-location normalization without biases yields a solution that satisfies the variance constraint---and we find that it consistently leads to stable training in all our experiments (discussed next). This highlights the importance of including our normalization strategy in the encoder network.

\begin{figure*}[!t]\centering
  \includegraphics[width=0.99\textwidth]{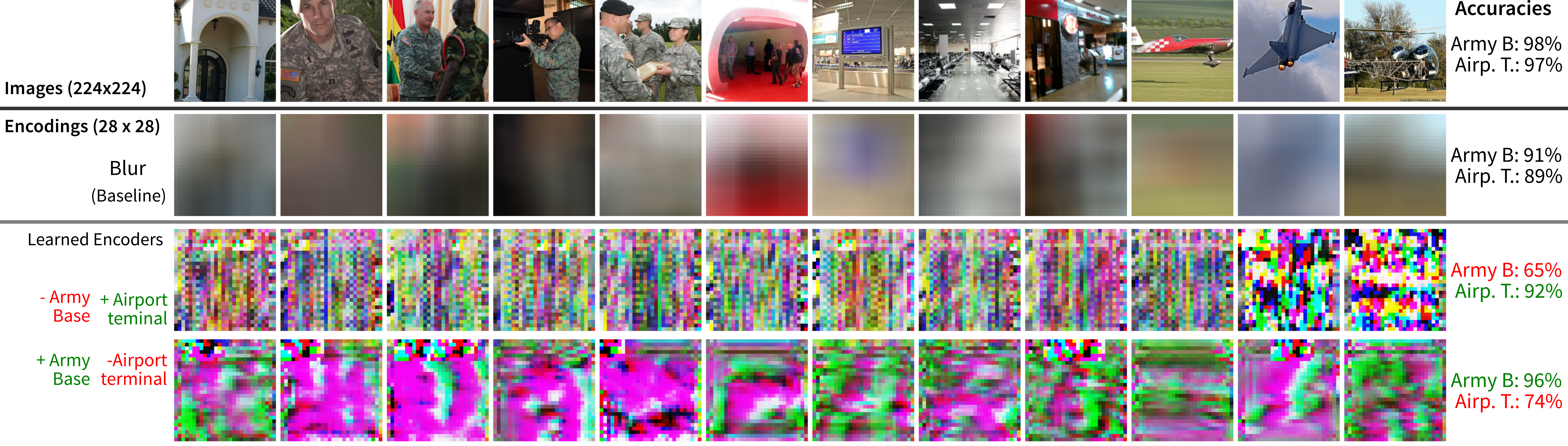}
  \caption{ \textbf{Visualization of Encoder Outputs.} We visualize outputs for various images from our learned encoders, with the far-right column indicating test accuracy for classifiers trained on encoded outputs. The first two rows show the original images, and blurred images we consider as a baseline. The remaining rows visualize encoders trained with our approach, to inhibiting (- sign in left column) and promote (+ sign in left column) specific scene detection tasks.}
\label{fig:viz}
\end{figure*}

\subsection{Evaluation of Privacy and  Utility}

Finally, we conduct a broad evaluation of our method's ability to inhibit private tasks while maintaining utility. We train encoders for different combinations of private tasks with both generic and task-specific utility constraints. For each encoder, we train classifiers using our verification protocol for a number of scene-detection tasks---tasks that the encoder was trained to (a) inhibit, (b) promote, or (c) neither to inhibit nor promote. We report the test set performance of these classifiers in Table~\ref{tab:results}, and for context, also report classifier performance on the original images themselves (reported as the ``Identity'' encoding), as well a simple blur baseline.  The blur baseline is not task specific, and simply produces $28\times 28$ image outputs by applying an $120\times 120$ averaging filter (our encoder's receptive field is $112\times 112$) and downsampling by a factor of $8$. While Table~\ref{tab:results} shows the final accuracy achieved by the classifiers, our encoders also slow down their training---Fig.~\ref{fig:teaser} illustrates this by showing the evolution of validation loss during classifier training. We also visualize some of the learned encoding functions in Fig.~\ref{fig:viz}, where we show examples of typically encoded images for each encoder. To better show the variability between images, we map the encoder output to an RGB image by mapping the value at each location and channel to a histogram equalized value.

We begin by discussing the performance of encoders trained against four group 1 tasks with only the generic variance constraint, and see that in every case, these cause considerable degradation in their corresponding private task accuracy over classification of original images, much more so than the blur baseline. Looking at the performance across tasks, it is apparent that some tasks are easier to solve and therefore harder to inhibit (e.g., see ``airfield''). This is likely because some categories have a larger number of redundant cues that are harder to effectively censor. This is why the learned encoders have different degrees of success in censoring different tasks. Interestingly, censoring such easily solved tasks leads to overall poorer performance for other tasks as well likely due to the encoder being forced to remove a lot of information that may also be useful for other tasks. Conversely, some tasks are hard to solve (like ``arch''), and these are easily inhibited even when they are not targeted by the encoder. But an encoder trained to inhibit these tasks is found to preserve classification accuracy for remaining tasks to a greater degree.

We next consider encoders that are trained using \eqref{eq:encdef2} to promote certain desirable tasks, while inhibiting the private task. This allows the encoder to retain specifically useful image cues, as opposed to simply preserving output variance. In Table~\ref{tab:results}, we find that this approach almost always yields high accuracy for the targeted desirable tasks, with little to no difference in the encoder's ability to inhibit the private task. Indeed, using this approach we are able to enable high accuracy for ``arch'' task (which suffered poor accuracy in all the generic variance encoders) while inhibiting ``army base''. Interestingly, preserving specific desirable tasks also has a generalization effect, with improved accuracy on tasks other than the desirable (and private) tasks. To this end, we train encoders with multiple desirable tasks (with $I_v$ formulated as an $N+1$-way classification for $N$ desirable tasks), and find that as we increase the set of positive tasks, the encoder generalizes to providing more and more general utility (albeit, with some degradation in privacy). This implies that by choosing a diverse set of positive tasks during training, an encoder can learn to retain information for a broad class of applications.

\section{Conclusion}
\label{sec:conc}

We presented an effective and practical method for learning image encoding functions that remove information related to sensitive private tasks. We considered a formulation based on adversarial optimization between the encoding function and estimators for the private tasks, modeling both as deep neural networks. We described a stable and convergent strategy for optimization, which yielded encodings that permanently inhibit recovery of private attributes while maintaining utility as shown experimentally with an exhaustive verification protocol.

Note that we did not constrain our encoded outputs to appear natural, or resemble the original data. Consequently, our framework requires classifiers for the desirable tasks to also be retrained. Others (e.g., \cite{meden2018k,brkic2017know,di2017gp}) have successfully incorporated  such requirements in different approaches to privacy and censorship, and one of our goals in future work is to extend our framework in a similar manner.

\noindent\textbf{Acknowledgments.} FP and SJK were supported by ONR N00014-18-1-2663, DHS 2014-DN-077-ARI083, and NSF 1514154. Views and conclusions in this paper are those of the authors, and do not necessarily represent official policies, either expressed or implied, of the US DHS.

{\small
\bibliographystyle{ieee}

}

\end{document}